\documentclass[letterpaper, 10 pt, conference]{ieeeconf}  

\IEEEoverridecommandlockouts                              

\usepackage{pifont}
\newcommand{\cmark}{\ding{51}}%
\newcommand{\xmark}{\ding{55}}%
\usepackage{graphics} 
\usepackage{epsfig} 
\usepackage{mathptmx} 
\usepackage[fleqn]{nccmath}
\usepackage{amssymb}  
\usepackage{subfig}
\usepackage{multirow}
\usepackage{booktabs}
\usepackage{footnote}
\usepackage[noadjust]{cite}
 
\usepackage{color} 
 
\title{\LARGE \bf
Design method for an anthropomorphic hand able to gesture and grasp}

\author{Giulio Cerruti$^{1}$, Damien Chablat$^{2}$, David Gouaillier$^{3}$ and Sophie Sakka$^{4}$
\thanks{$^{1}$Giulio Cerruti is with Institut de Recherche en Communications et Cybern\'etique de Nantes (IRCCyN), 1 rue de la No\"e BP 92101, 44321 Nantes Cedex 3, and Aldebaran in the Advanced Mechatronic Lab (AM-Lab Nantes), 6 rue Saint Domingue, 44200 Nantes, France\newline {\tt gcerruti@aldebaran.com} }%
\thanks{$^{2}$D. Chablat is with the LUNAM University, IRCCyN UMR CNRS 6597, France {\tt Damien.Chablat@irccyn.ec-nantes.fr}}%
\thanks{$^{3}$D. Gouaillier is head of Aldebaran AM-Lab Nantes, 44200 Nantes, France {\tt dgouaillier@aldebaran.com} }
\thanks{$^{4}$S. Sakka is with the LUNAM University, IRCCyN UMR CNRS 6597, and University of Poitiers, 93 Avenue du Recteur Pineau, 86000 Poitiers, France {\tt Sophie.Sakka@irccyn.ec-nantes.fr}}
}%

\begin{document}

\maketitle
\thispagestyle{empty}
\pagestyle{empty}

\begin{abstract}
This paper presents a numerical method to conceive and design the kinematic model of an anthropomorphic robotic hand used for gesturing and grasping. In literature, there are few numerical methods for the finger placement of human-inspired robotic hands. In particular, there are no numerical methods, for the thumb placement, that aim to improve the hand dexterity and grasping capabilities by keeping the hand design close to the human one. While existing models are usually the result of successive parameter adjustments, the proposed method determines the fingers placements by mean of empirical tests. Moreover, a surgery test and the workspace analysis of the whole hand are used to find the best thumb position and orientation according to the hand kinematics and structure. The result is validated through simulation where it is checked that the hand looks well balanced and that it meets our constraints and needs. The presented method provides a numerical tool which allows the easy computation of finger and thumb geometries and base placements for a human-like dexterous robotic hand.\\
\noindent
\textit{Keywords: robotic hand, gesture, hand dexterity, Kapandji test, thumb opposability, grasping.}   
\end{abstract}

\section{Introduction}
\label{sec:Introduction}
The human hand is an astonishingly advanced mechanism which is too complicated to be faithfully replicated. Many techniques are used to model its kinematics, such as direct measurements of the human limbs or optimization algorithms. Currently, the most accurate hand model ever built is the anatomically-correct testbed (ACT) \cite{ACT}. In this design, the structure was machined with the same shape and mass to that of the human bones, and the joints were designed to preserve the same DoFs and passive stiffness of the human joints. However, the ACT is conceived to deeply investigate the human hand structure, function and control for medical purposes. Its complexity hinders the practical use of the artifact in robotics. In general, simplified joint structures and link shapes are commonly chosen by robotic hand designers. Furthermore, the number of independent DoFs are usually reduced due to technological limits (hardware and software) and mechanical constraints.\\
Each human hand has its own peculiarities, and yet, its functional capabilities do not significantly change from the others. This implies that different link lengths, width and joint placements do not greatly alter the hand performance. As a consequence, no true optimal design exists for the construction of a human-like robotic hand. Nevertheless, an appropriate model has to be realized in order to fulfill all functional requirements: the hand is supposed to be able to express emotions and give information through gestures, grasping objects and manipulating them. These abilities require high interaction among fingers, in particular between each finger and the thumb. Therefore, finger geometry and kinematics are less important than the thumb kinematics and its interaction with the opposed fingers.\\
Our project aims to design a self-contained human-inspired hand with following constraints: have the size and weight similar to a 6 years old child's hand (length approximately 120 mm and weight less than 0.6 Kg) in order to be embodied on a humanoid robot of a comparable height, be able to grasp small objects (such as a full soda can) and perform gestural communication (e.g. thumb-up, ok, pointing and counting). This document is the first step in the design process for which we solve geometric problems only. Section \ref{sec:hi conception} presents a literature review on human hands and existing robotic hands. It will be shown that no hand that meets our constraints exists. Sections \ref{sec:hand design} addresses the problem presenting the kinematic aspects of our hand. Section \ref{sec:thumb base selection} proposes a new design method to place the thumb base in order to fulfill the required functional capabilities. Finally, results are discussed in Section \ref{sec:validation and discussion} and conclusions are drawn in Section \ref{sec:Conclusions}.



\section{Literature Review}
\label{sec:hi conception}
\subsection{The human hand}
\label{subsec:human hand}
\subsubsection{Hand skeleton}
\begin{figure}[tb]
\centering
\includegraphics[scale = 0.14]{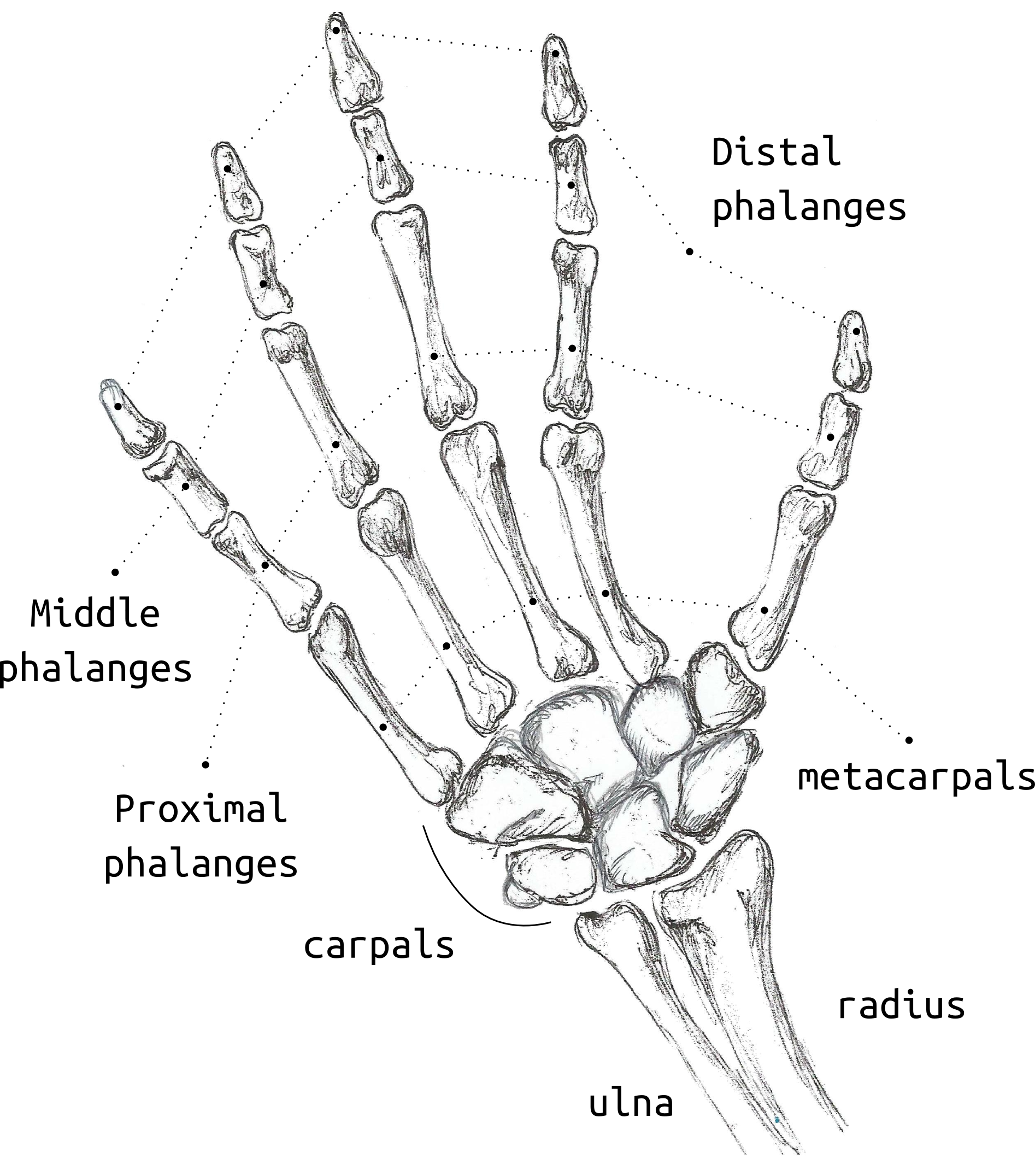}
\caption{Drawing of the human hand skeleton.}
\label{fig:handSkeleton}
\end{figure}

The human hand is composed of 27 bones \cite{HumanHandModel} which make up its three main parts: wrist, palm and fingers (Figure \ref{fig:handSkeleton}). The wrist is formed by 8 small bones called carpals which join the ulna and radius forearm bones to the hand. The palm is composed by 5 bones called metacarpals which connect the fingers and the thumb to the wrist. The joints between the wrist and the finger metacarpals are called the carpometacarpal (CMC) joints while the one linking the metacarpal of the thumb is called the radiocarpal (RC) joint. Each finger is composed by 3 long bones called phalanges whose names are given according to their distance to the palm: proximal, middle and distal. The proximal phalanx (PP) is linked to the metacarpal bone through the metacarpophalangeal joint (MCP) while the remaining phalanges are connected to each other through the interphalangeal joints. The joint between the proximal and the middle phalanx is the proximal interphalangeal joint (PIP) while the joint closest to the end of the finger is the distal interphalangeal joint (DIP). The thumb has one bone less with respect to the fingers (no middle phalanx) and it consequently has one interphalangeal joint plus the MCP joint.

\subsubsection{Hand kinematics}
\begin{figure}[tb]
\centering
\includegraphics[scale = 0.53]{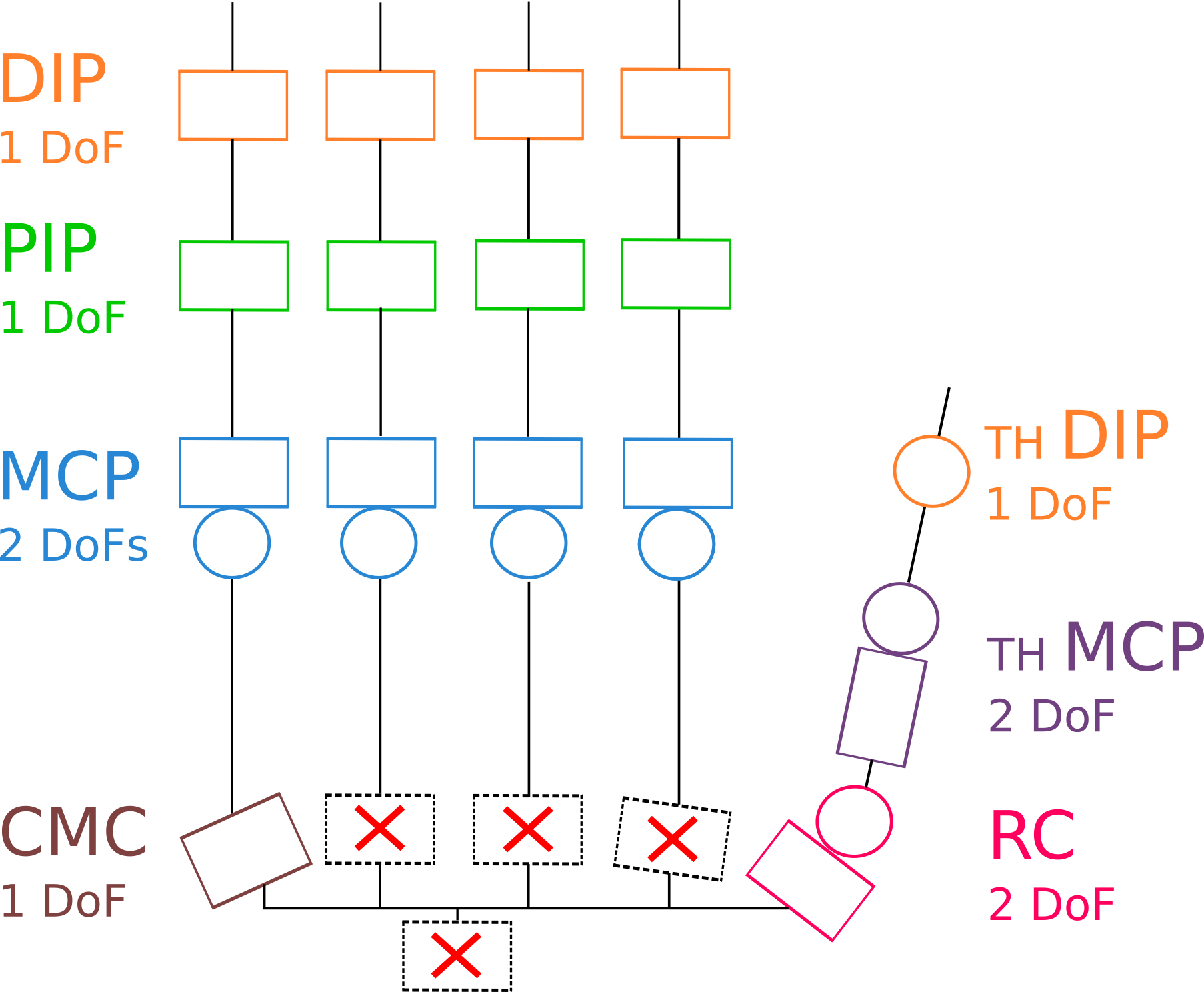}
\caption{Top view of simplified human hand kinematic model (22 DoFs). Note that the drawing does not aim to reproduce the real structure.}
\label{fig:handKinematics}
\end{figure}

The hand joints are classified into three main types: hinge (1 DoF), condyloid (2 DoFs) and saddle (2 DoFs) joints. The hinge joints on the human hand are the CMC, PIP and the DIP joints. They allow the fingers to flex (move toward the palm) and to extend (move further from palm). The condyloid joints are the finger MCP joints that allow flextion/extension and abduction/adduction of the fingers, the latter being the motion of spreading and gathering them. Hence, each finger can be represented as a kinematic serial chain of 5 DoFs: 1 at the CMC, 2 at the MCP, 1 at the PIP and a last one at the DIP. Finger PIP and DIP joints are orthogonal to the bone axis when the phalanx is fully extended and they progressively bend toward the middle (due to the bone surface) while flexing. As a result, all fingers converge to a common point improving the opposition of the thumb to the ring and little fingers. According to Vitruvian man's hand and the study conducted by Isobe \cite{IsobeHandPosition}, human fingertips approximatively lie on a common circle when abducted. The circle has a radius equal to the middle finger length and it is centered at the MCP joint of the middle finger. The thumb has the same number of DoFs of fingers but differently distributed: the RC joint is a saddle joint, the MCP is a condyloid joint and the IP joint is a hinge joint \cite{fiveLinkModelThumb}. The thumb is the only finger able to turn and oppose to the other four fingers. The opposability of the thumb enables humans to grip and hold objects that they would not be able to take otherwise. Figure \ref{fig:handKinematics} shows the summarized DoFs of the human hand plus a further DoF in the palm that represents the small motions that occur among the carpal bones while flexing the wrist. However, some joints are relatively immobile due to the interousseous ligaments that stabilize the hand joints. The CMC joints at the bases of the index, middle and ring metacarpals can be neglected and the small intra-carpal motions can be collected into the wrist motion. Note that the CMC joint of the little finger should not be omitted since it forms the hollowed shape of the palm \cite{jointStructure} when the little finger moves in opposition to the thumb. Not all DoFs in the human hand are independent. Tendons couple some joint like the PIP and DIP joints of the fingers. In the next paragraphs we will refer to the independent DoFs simply as DoFs.

\subsection{Robotic hands}
\label{subsec:robotis hand kinematics}
The number of joints and DoFs of human-inspired robotic hands is generally chosen according to the tasks for which they are conceived. For instance, robotic hands oriented to reproduce the human hand manipulation capabilities adopt nearly the same kinematic structure presented in the previous paragraph. Excluding the wrist, the Shadow Dexterous Hand \cite{ShadowHand} has 18 independent DoFs, 5 of which are dedicated to the thumb, with the PIP and DIP joints coupled. The Awiwi hand, mounted on the DLR Hand Arm System \cite{DLRhandArmSystem}, has even more independent DoFs (19): its PIP and DIP joints of the index and middle are not coupled and it has a simplified thumb with 4 joints and 4 DoFs. The UB Hand IV \cite{UBhandIV} and DLR/HIT Hand II \cite{DLRHIThandII} (15 DoFs) do not include the CMC joint of the little finger in their kinematic models and their thumbs have 4 joints with the MCP and the IP coupled.\\ Robotic hands designed only for grasping require less complex kinematic models with less joints and DoFs. Generally they have all finger joints coupled, the CMC joint of the little finger coupled with the thumb base motion and fingers with no abduction/adduction motion (MCP with 1 DoF). The Harada  Robot Hand \cite{HaradaHand}, for instance, has only 14 joints (3 per finger and 2 for the thumb) and 5 DoFs: 1 DoF per finger and 1 DoF for the thumb. Another example is the Prensilia IH2 Azzurra hand \cite{AzzurraHand} that has 11 joints, 5 of which actuated: 2 DoFs for the thumb, 1 DoF for index and middle and the last DoF for both ring and little fingers. Robotic hands that aim to both manipulate and grasp find a compromise between hand motion capabilities and control complexity. Among these we find hands such as the Robonaut II Hand \cite{RobonautIIhand}, Elu2-Hand \cite{EluHand} (Shunk hand) and the iCub hand \cite{iCubHand}. Table \ref{tab:RoboticHandsComparison} 
lists the DoFs and the weight of the aforementioned hands and it specifies if hands fully embed their actuators and electronic components. As it can be seen, none of them fulfill our requirements: half of them are not self-contained (i.e. they have hardware placed externally from the hand), the DLR/HIT Hand II is too heavy, the Harada Robot Hand, Prensilia IH2 Azzurra hand and TUAT/Karlsruhe Hand \cite{TUATKarlsruheHand} do not have enough DoFs to perform the required gestures. Only the Elu2-Hand could fulfill our functional needs, however, its size is similar to that of an adult's hand and it cannot be scaled to the demanded size.
\begin{table}[tb]
\centering
\caption{Comparison of robotic hands. Wrist not included in the considered DoFs.}
\begin{tabular}{cccc}
\multicolumn{1}{c}{\multirow{1}{*}{\textbf{Name}}} & \multicolumn{1}{c}{\multirow{1}{*}{\textbf{DoFs}}} & \multicolumn{1}{c}{\multirow{1}{*}{\textbf{Weight [Kg]}}} & \multicolumn{1}{c}{\multirow{1}{*}{\textbf{Self-contained}}}\\
\midrule
\multicolumn{1}{c}{DLR Hand Arm System} & \multicolumn{1}{c}{19} & \multicolumn{1}{c}{\multirow{1}{*}{13.5\textsuperscript{*}}} & \multicolumn{1}{c}{\multirow{1}{*}{\xmark}}\\
\multicolumn{1}{c}{Shadow Dexterous Hand} & \multicolumn{1}{c}{18} & \multicolumn{1}{c}{\multirow{1}{*}{4.2}} & \multicolumn{1}{c}{\multirow{1}{*}{\xmark}}\\
\multicolumn{1}{c}{UB Hand IV} & \multicolumn{1}{c}{15} & \multicolumn{1}{c}{\multirow{1}{*}{n.d.}} & \multicolumn{1}{c}{\multirow{1}{*}{\xmark}}\\
\multicolumn{1}{c}{DLR/HIT Hand II} & \multicolumn{1}{c}{15} & \multicolumn{1}{c}{\multirow{1}{*}{1.5}} & \multicolumn{1}{c}{\multirow{1}{*}{\cmark}}\\
\multicolumn{1}{c}{Robonaut 2 Hand} & \multicolumn{1}{c}{12} & \multicolumn{1}{c}{\multirow{1}{*}{2.25}} & \multicolumn{1}{c}{\multirow{1}{*}{\xmark}}\\
\multicolumn{1}{c}{Elu2-Hand} & \multicolumn{1}{c}{9} & \multicolumn{1}{c}{\multirow{1}{*}{0.74}} & \multicolumn{1}{c}{\multirow{1}{*}{\cmark}}\\
\multicolumn{1}{c}{iCub hand} & \multicolumn{1}{c}{9} & \multicolumn{1}{c}{\multirow{1}{*}{0.212}} & \multicolumn{1}{c}{\multirow{1}{*}{\xmark}}\\
\multicolumn{1}{c}{Harada Robot Hand} & \multicolumn{1}{c}{5} & \multicolumn{1}{c}{\multirow{1}{*}{0.369}} & \multicolumn{1}{c}{\multirow{1}{*}{\cmark}}\\
\multicolumn{1}{c}{IH2 Azzurra} & \multicolumn{1}{c}{5} & \multicolumn{1}{c}{\multirow{1}{*}{0.64}} & \multicolumn{1}{c}{\multirow{1}{*}{\cmark}}\\
\multicolumn{1}{c}{TUAT/Karlsruhe Hand} & \multicolumn{1}{c}{1} & \multicolumn{1}{c}{\multirow{1}{*}{0.125}} & \multicolumn{1}{c}{\multirow{1}{*}{\cmark}}\\
\midrule
\multicolumn{4}{l}{\textsuperscript{*}\footnotesize{considering arm and hand together.}}
\end{tabular}
\label{tab:RoboticHandsComparison}
\end{table}

\noindent
In general, anthropomorphic robotic hands kinematics are not designed on the base of the finger-thumb interaction, even though hand performance strictly depends on the thumb dexterity and opposability. Thumb capabilities are determined by its kinematics and base placement. The problem of thumb placement has already been addressed in literature. Grebenstein \textit{et al.} \cite{7BillionPerfectHands} developed cardboard prototypes to incrementally refine the hand kinematics in order to obtain thumb functionalities and hand aesthetics similar to the human one. Wang \textit{et al.} \cite{designGuidelineForTheThumb} approached the problem using a numerical method based on the Euler rotation theorem. The thumb placement was computed passing from a lateral posture to an opposing one in order to ensure basic grasping modes. This paper solves the problem using a numerical method based on a surgery test and a workspace analysis of the whole hand. The best thumb position and orientation is found according to the hand kinematics and structure.

\section{Our hand design}
\label{sec:hand design}
The design of the kinematic model of our hand started from mechanical and functional observations which lead to the following objectives:
\begin{itemize}
\item be aesthetically anthropomorphic in order to be easily accepted by human beings;
\item be proportioned to a 6 years old child's hand;
\item weigh less than 0.6 Kg - to be embodied on the robot;
\item be compliant - to have a safe human-robot interaction;
\item be silent.
\end{itemize}
while from a functional point of view it has to:
\begin{itemize}
\item 70$\%$ communicate information and emotions using gestures: pointing, counting, thumb-up, ok, etc \ldots;
\item 30$\%$ grasp light objects, like a smart-phone or a full soda can.
\end{itemize}

\subsection{Kinematics design}
\label{subsec:Our kinematics design}
\begin{figure}[tb]
\centering
\includegraphics[scale = 0.5]{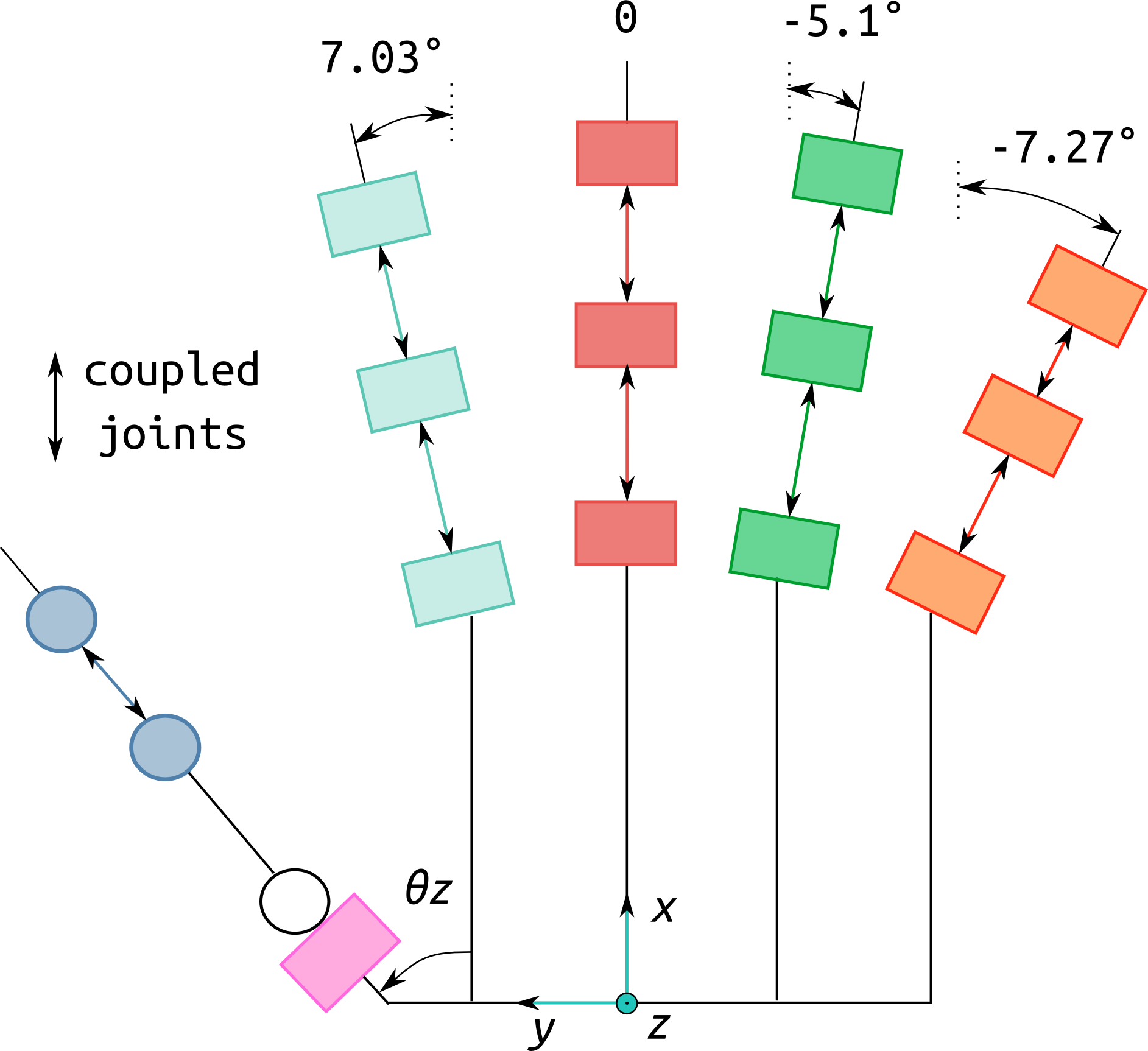}
\caption{Simplified kinematic model of our robotic hand. Each color refers to a single DoF.}
\label{fig:firstImplemendtedModelPaper}
\end{figure}

To have a light and small hand the number of actuators has to be as small as possible. A robotic hand that is able to grasp various objects does not need numerous actuators. Thanks to differential mechanisms and under-actuated fingers capable to adapt to different surfaces \cite{Berglin} only one actuator is sufficient \cite{TUATKarlsruheHand}. Also finger geometry and couplings based on postural synergies \cite{BicchiSynergies} allow to use a reduced number of actuators. However, in our case, non-verbal communication should be added to grasping capabilities. For this reason, the design of the kinematic model is firstly oriented to satisfy the required gestures. To achieve this, the whole human hand kinematic model is firstly considered with all its joints coupled. Then, for each gesture, joint couplings are incrementally relaxed to minimally increase the number of actuators. After that, the number of joints is reduced to meet the weight and size limits. Obviously, this last step imposes a trade off between well-mimicked gestures and a light and small structure. To identify the less relevant joints, gestures are classified according to the feelings they trigger in the recipient: low priority is given to gestures inducing negative sensations, while, middle and high priorities are associated to the ones inspiring neutral and positive feelings respectively. Hence, to meet weight and size constraints, joints purely involved in performing low priority gesture are removed.
The thumb is defined with four joints, abduction/adduction finger motions are removed as well as the CMC joint at the base of the little finger. Figure \ref{fig:firstImplemendtedModelPaper} shows our final hand kinematic model: 1 DoF per finger (they can only flex/extend) and 3 DoFs for the thumb. All joints can rotate by 90$^{\circ}$ about their axes. Their range of motion normally goes from 0$^{\circ}$ to 90$^{\circ}$, where 0$^{\circ}$ refers to the hand configuration in which all fingers are fully extended on the same plane.


\subsection{Finger and thumb proportions} 
\label{subsec:finger and thumb proportions}
No best practice to define finger lengths, widths and placements exists. Modular designs of some or all fingers are often used to help mechanical modeling and construction. Each finger has the same link length, width, joint RoM and DoFs and their placement on the palm determines the similarity to the human counterpart. Nevertheless, a better affinity with the human hand is achieved respecting human finger diversities. Link lengths and widths are derived from body proportions under the assumption that the structure and bones proportions of hands are maintained despite their physical size \cite{basicResearchHandBiomechanics}. Link lengths are commonly disclosed as percentage of the hand length \cite{Davidoff}, therefore, they can be easily computed knowing the distance between the wrist and the middle fingertip. Link widths measurements are determined as a function of the hand width using two linear equations as lower and upper bounds. This approach took inspiration from the comparison done by \cite{fingerWidths} between the Grainer model \cite{greinerHand} and Buchholz and Armstrong model \cite{BuchholzHand}. 
Table \ref{tab:FingerProportions} gives the link lengths and widths for a humanoid robot of similar height to a 6 year old child.
\begin{table}[tb]
\centering
\caption{Link lengths and widths for a humanoid robot of similar height to a 6 year old child.}
\begin{tabular}{cccc}
\multicolumn{1}{c}{\multirow{1}{*}{\textbf{Finger}}} & 
\multicolumn{1}{c}{\multirow{1}{*}{\textbf{Link}}} & 
\multicolumn{1}{c}{\multirow{1}{*}{\textbf{Length [mm]}}} & \multicolumn{1}{c}{\multirow{1}{*}{\textbf{Width [mm]}}}\\
\midrule
\multicolumn{1}{c}{\multirow{3}[2]{*}{\textbf{Thumb}}} & \multicolumn{1}{c}{Carpometacarpal} & \multicolumn{1}{c}{30.8} & \multicolumn{1}{c}{25}\\
\multicolumn{1}{c}{} & \multicolumn{1}{c}{Proximal phx} & \multicolumn{1}{c}{22.6} & \multicolumn{1}{c}{16}\\
\multicolumn{1}{c}{} & \multicolumn{1}{c}{Distal phx} &
\multicolumn{1}{c}{20.7} & \multicolumn{1}{c}{16}\\
\midrule
\multicolumn{1}{c}{\multirow{3}[2]{*}{\textbf{Index}}} & \multicolumn{1}{c}{Proximal phx} & \multicolumn{1}{c}{26.4} &
\multicolumn{1}{c}{15}\\
\multicolumn{1}{c}{} & \multicolumn{1}{c}{Middle phx} & 
\multicolumn{1}{c}{17.1} & 
\multicolumn{1}{c}{14}\\
\multicolumn{1}{c}{} & \multicolumn{1}{c}{Distal phx} &
\multicolumn{1}{c}{16.8} &  \multicolumn{1}{c}{13}\\
\midrule
\multicolumn{1}{c}{\multirow{3}[2]{*}{\textbf{Middle}}} & \multicolumn{1}{c}{Proximal phx} & \multicolumn{1}{c}{29.7} & \multicolumn{1}{c}{15}\\
\multicolumn{1}{c}{} & \multicolumn{1}{c}{Middle phx} &
\multicolumn{1}{c}{19.1} & \multicolumn{1}{c}{14}\\
\multicolumn{1}{c}{} & \multicolumn{1}{c}{Distal phx} & 
\multicolumn{1}{c}{18.2} & 
\multicolumn{1}{c}{13}\\
\midrule
\multicolumn{1}{c}{\multirow{3}[2]{*}{\textbf{Ring}}} & \multicolumn{1}{c}{Proximal phx} & \multicolumn{1}{c}{26.9} & \multicolumn{1}{c}{15}\\
\multicolumn{1}{c}{} & \multicolumn{1}{c}{Middle phx} &
\multicolumn{1}{c}{18.5} &  \multicolumn{1}{c}{14}\\
\multicolumn{1}{c}{} & \multicolumn{1}{c}{Distal phx} &
\multicolumn{1}{c}{18.1} &  \multicolumn{1}{c}{13}\\
\midrule
\multicolumn{1}{c}{\multirow{3}[2]{*}{\textbf{Little}}} & \multicolumn{1}{c}{Proximal phx} & \multicolumn{1}{c}{21.4} & \multicolumn{1}{c}{13}\\
\multicolumn{1}{c}{} & \multicolumn{1}{c}{Middle phx} &
\multicolumn{1}{c}{13.1} &  \multicolumn{1}{c}{12}\\
\multicolumn{1}{c}{} & \multicolumn{1}{c}{Distal phx} &
\multicolumn{1}{c}{15.7} &  \multicolumn{1}{c}{11}\\
\midrule
\end{tabular}
\label{tab:FingerProportions}
\end{table} 

\subsection{Finger placement} 
\label{subsec:finger placement}
\begin{figure}[tb]
\centering
\includegraphics[scale = 0.35]{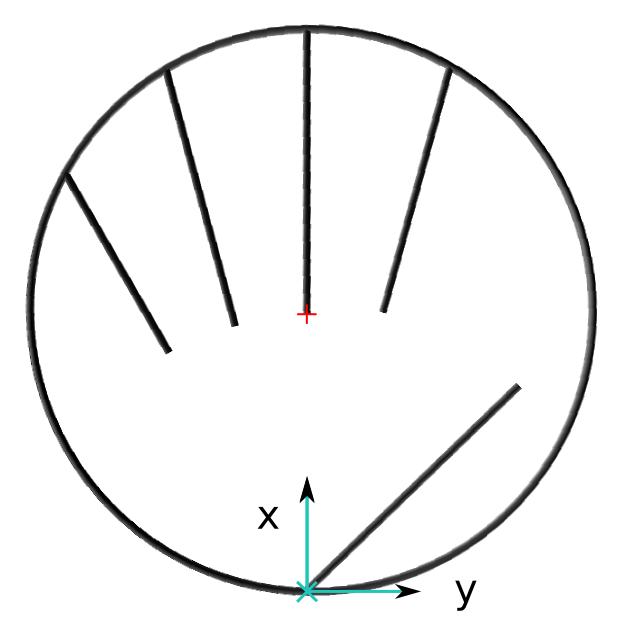}
\caption{Finger base placement approach.}
\label{fig:15DegFingerPos}
\end{figure}

Finger bases are computed using a circle centered at the MCP joint of the middle finger \cite{IsobeHandPosition}. Since link lengths are fixed (Table \ref{tab:FingerProportions}), the angle between each finger base determines the linking point to the palm. Small angles return closer finger bases, with similar heights from the wrist, while big angles locate finger bases at further distances and with various heights from the wrist. Obviously, too close finger bases are not mechanically feasible, while far finger bases result in an unaesthetic design. A compromise between the two is necessary in order to have a well-balanced hand (Figure \ref{fig:15DegFingerPos}). Along the $z$ axis, fingers are placed so that the fingertips lie on a common $xy$ plane when they are completely flexed. This allows the palm to have the same arched shape as the human hand.\\
Designed fingers can only flex and extend (1 DoF) via three parallel hinge joints. 
Hence, MCP joints are inclined to preserve the finger convergence of the human hand towards the center of the palm. To determine maximum angles of inclination of the finger bases, intra-finger collisions are checked while flexing the joints. 
For each finger, finger distances are computed increasing the inclination angle from $0^\circ$ until a collision is detected (Figure \ref{fig:firstImplemendtedModelPaper}). Table \ref{tab:Baseplacement} shows the resulting finger base positions and orientations. The distance along y between the little and the ring bases is smaller than the others, slightly complicating the mechanical design. However, with these results the aesthetics of the human hand is globally met.

\begin{table}[tb]
\centering
\caption{Finger base positions and orientations with respect to the hand frame (Figure \ref{fig:firstImplemendtedModelPaper}).}
\begin{tabular}{ccc}
\multicolumn{1}{c}{\multirow{1}{*}{\textbf{Finger}}} & \multicolumn{1}{c}{\multirow{1}{*}{\textbf{Base position [mm] ($x$, $y$, $z$)}}} & \multicolumn{1}{c}{\multirow{1}{*}{\textbf{Base orientation [deg]}}}\\
\midrule
\multicolumn{1}{c}{\multirow{1}{*}{\textbf{Index}}} & \multicolumn{1}{c}{(59.3, 18.1, -1.8)} & \multicolumn{1}{c}{7.03$^{\circ}$}\\
\midrule
\multicolumn{1}{c}{\multirow{1}{*}{\textbf{Middle}}} & \multicolumn{1}{c}{(60.5, 0.0, 0.0)} & \multicolumn{1}{c}{0$^{\circ}$}\\
\midrule
\multicolumn{1}{c}{\multirow{1}{*}{\textbf{Ring}}} & \multicolumn{1}{c}{(56.1, -17.2, -2.7)} & \multicolumn{1}{c}{-5.1$^{\circ}$}\\
\midrule
\multicolumn{1}{c}{\multirow{1}{*}{\textbf{Little}}} & \multicolumn{1}{c}{(46.1, -31.1, -6.8)} & \multicolumn{1}{c}{-7.27$^{\circ}$}\\
\midrule
\end{tabular}
\label{tab:Baseplacement}
\end{table}

\subsection{Thumb placement}
\label{subsec:thumb placement}
To approximately attain human hand performances the thumb geometry and kinematics need to be carefully designed. To gesture and grasp, 4 joints with 3 DoFs are chosen for the thumb kinematics (Figure \ref{fig:firstImplemendtedModelPaper}). Its proportions are calculated as described in the previous Section \ref{subsec:finger and thumb proportions}. The thumb base configuration is computed following three criteria: dexterity, opposability and aesthetics. In order to imitate human thumb motions, the robotic thumb has to be able to reach a certain number of desired positions on the opposite fingers. These positions are defined by a well-known surgery test, known as Kapandji test \cite{Kapandji}, used after pollicization, which consists in reconstructing the thumb by displacing the index finger from its MCP joint to the MCP joint of the thumb. The test contains all motion directions and an interesting set of positions which are useful to check the thumb range of motion and its dexterity. However, it cannot provide information about thumb grasping abilities. Reaching all required positions does not necessarily imply that the thumb is effectively able to interact with the opposed finger. Hence, an additional test is carried out to explore the physical interaction between the thumb fingertip and the opposed fingers. This test returns the intersection volume between the thumb and fingers as a thumb opposability index \cite{GifuIII}, which denotes the hand grasping abilities from a kinematics point of view. Note that no forces are considered in this context; an additional evaluation metric should be taken into account to determine finger forces and grasping stability. Finally, the global appearance of the hand has to be equilibrated and easily accepted. Indeed, the hand will be mounted on a social robot and its aesthetics is fundamental to enhance and encourage human-robot interaction.\\
In summary, the thumb base placement is the most delicate design step. In order to obtain the needed functionalities (gesturing and grasping) the thumb has to be dexterous and opposable. These two properties are respectively achieved through:
\begin{itemize}
\item The Kapandji test;
\item The thumb opposability index.
\end{itemize}
In addition, an aesthetic check is done to evaluate the overall hand appearance. This last test implies a cyclic design in which the intervals of search among the thumb base parameters are adapted at each iteration.

\section{Thumb base selection}
\label{sec:thumb base selection}
In this section the tests used to design a dexterous and opposable thumb are presented. According to the thumb geometry and kinematics, the thumb base configuration is determined by calculating the thumb base position ($x$, $y$, $z$) and orientation about the $z$ ($\theta_z$) axis with respect to the hand frame (Figure \ref{fig:firstImplemendtedModelPaper}). The orientations about $x$ and $y$ axes are not included because of the 2 DoFs of the base.

\subsection{The Kapandji test}
\label{subsec:kapandji test}
\begin{figure}[tb]
\centering
\includegraphics[scale = 0.35]{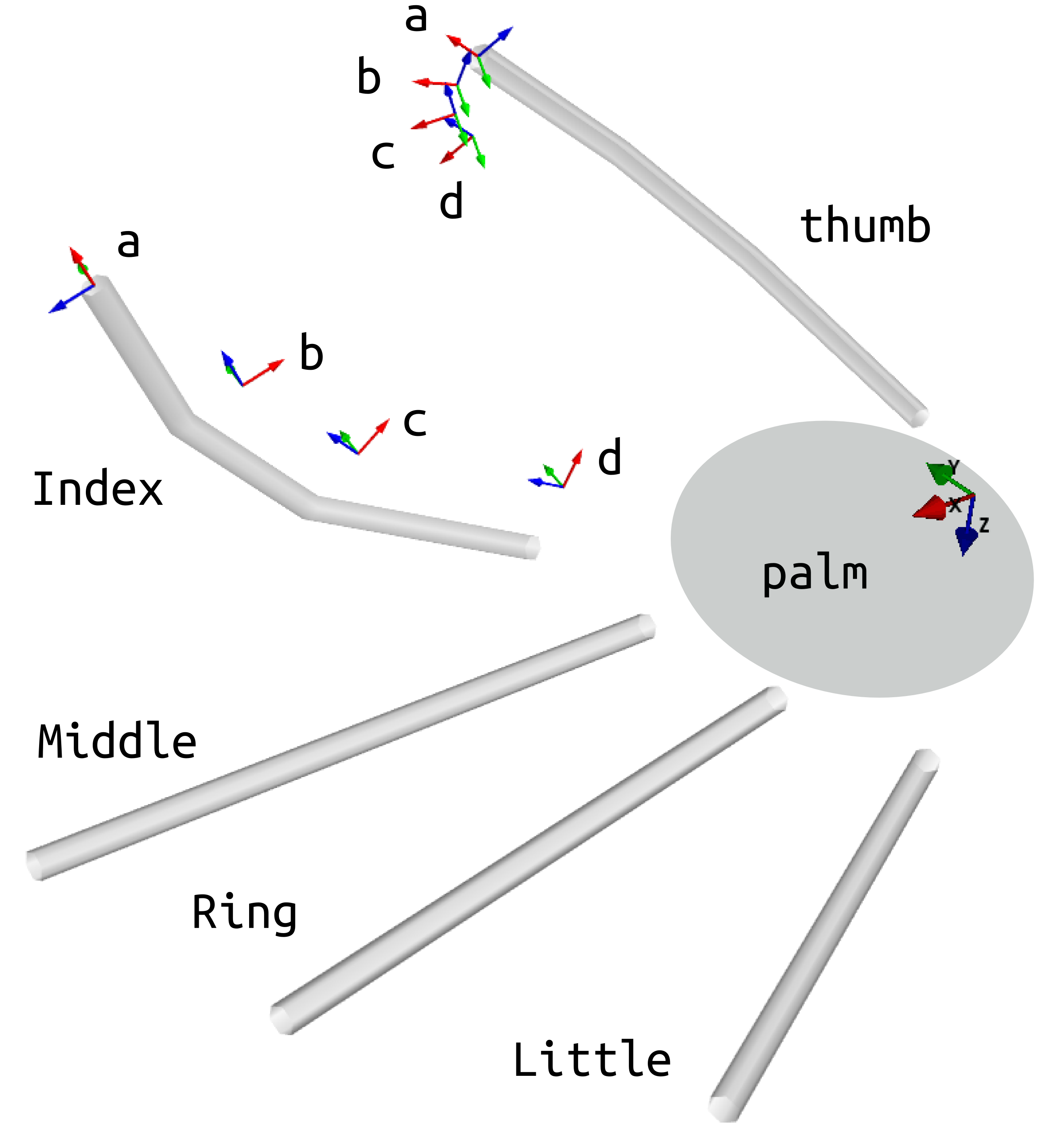}
\caption{Desired positions defined at the joints and fingertip of the opposing finger. Target frames are placed exactly on the finger surface. The image shows simplified finger links to facilitate the view. As it can be seen, for each desired position on the finger (a, b, c, d), the thumb end-effector at the pulp is accordingly displaced so that it will reach the desired positions in configurations similar to the ones of the human hand.}
\label{fig:problemDefinitionForOneFinger}
\end{figure}

The Kapandji test assigns a score for each reached position of the thumb on the opposite fingers. As shown in \cite{Kapandji} the thumb passes from the index base to its fingertip and proceeds to the little fingertip touching the end of the fingers in between. Finally, it reaches the little base to complete the test. Based on this idea, the thumb is asked to reach all finger joints and fingertips for a total of 16 desired positions. If the thumb attains all 4 positions defined on each finger, the thumb base is collected as a candidate solution (Figure \ref{fig:problemDefinitionForOneFinger}).
 
\subsubsection{Problem formulation}
\label{subsubsec: problem formulation}
The thumb and the finger in opposition are treated as two independent serial chains attached to a common base. On each chain an end-effector is defined according to the desired position that has to be reached. The objective is to determine if both kinematic chains can converge to a common position.\\

\noindent
\textit{Definitions:}
\begin{itemize}
\item $j = t$ or $f_i$ respectively referring to the thumb fingertip and the $i^{th}$ end-effector ($i =$ MCP, PIP, DIP and fingertip) on the $f^{th}$ finger;
\item $\mathbf{q}_j$ vector of the independent joint variables of the $j^{th}$ chain;
\item $T_j(\mathbf{q}_j)$ direct geometric model (DGM) of the $j^{th}$ serial kinematic chain. It computes the end-effector configuration passing from the joint space to the Cartesian space;
\item $J_j(\mathbf{q}_j)$ Jacobian matrix. It provides the $j^{th}$ end-effector differential displacement (linear and angular) in terms of the differential variation of the joint variables. In this context, only its upper part is used in order to uniquely consider the linear displacement of the end-effector: $\mathbf{dP} = J_j(\mathbf{q}_j)\mathbf{dq}_j$. 
\end{itemize}
The problem can be solved using the inverse geometric model where the desired frame changes at each iteration:
\begin{enumerate}
\item Initialization of current joint variables $\mathbf{q}^c_t$ and $\mathbf{q}^c_{f_i}$ to a random value or desired initial value within the joint domain;
\item Computation of the current end-effector transforms $T_t(\mathbf{q}^c_t)$ and $T_{f_i}(\mathbf{q}^c_{f_i})$ of each chain (thumb and opposing finger);
\item Computation of the position error defined as: \\ $\mathbf{dP} = P_t(\mathbf{q}^c_t) - P_{f_i}(\mathbf{q}^c_{f_i})$, where $P_t(\mathbf{q}^c_t)$ and $P_{f_i}(\mathbf{q}^c_{f_i})$ are the position vectors of $T_t(\mathbf{q}^c_t)$ and $T_{f_i}(\mathbf{q}^c_{f_i})$ respectively;
\item If ($||\mathbf{dP}||$ is as small as required): the algorithm stops;\\
else: $\mathbf{dP}$ is scaled if it is too big to be used in the differential model $\mathbf{dP} = \mathbf{dP}/|| \mathbf{dP} || dx$, where $dx$ is a properly defined small displacement;
\item Computation of thumb and finger current Jacobians $J_t(\mathbf{q}^c_t)$ and $J_{f_i}(\mathbf{q}^c_{f_i})$;
\item Calculation of joint variations $\mathbf{dq}_t = J_t(\mathbf{q}^c_t)^+ \mathbf{dP}/2$ and $\mathbf{dq}_{f_i} = J_{f_i}(\mathbf{q}^c_{f_i})^+ (-\mathbf{dP}/2)$;
\item Updating joint configurations $\mathbf{q}^c_t = \mathbf{q}^c_t + \mathbf{dq}_t$ and $\mathbf{q}^c_{f_i} = \mathbf{q}^c_{f_i} + \mathbf{dq}_{f_i}$ within the joint domain;
\item Reiteration from step 2.
\end{enumerate}
In our context, $\mathbf{q}_{f_i}$ is always a scalar for each finger $f$, independently from the chosen end-effector since all finger joints are coupled. Finger joint couplings influence the thumb base selection since end-effector trajectories change. In this context, transmission ratios are set to 1:1. Note that, if more than 3 DoFs are available on a single chain, this algorithm can be extended when computing the joint displacements. Indeed, a correction in orientation, if desired, could be attained as a secondary objective, working in the null space of the Jacobian matrix of the serial chain.
This iterative approach can return some undesired solutions: the thumb end-effector reaches the desired position passing through the finger in opposition. Consequently, a non-linear optimization algorithm with collision avoidance constraints is used: 
\begin{equation}
\begin{aligned}
&\min_{\mathbf{q}_t,\mathbf{q}_{f_i}} ||\mathbf{dP(\mathbf{q}_t,\mathbf{q}_{f_i})}||\\
\text{s.t.}&\\
& q_{t_l \min} \leq q_{t_l} \leq q_{t_l \max}, & l = 1, \text{ \ldots, DoFs}(\mathbf{q}_t)\\
& q_{{f_i}_l \min} \leq q_{{f_i}_l} \leq q_{{f_i}_l \max}, & l = 1, \text{ \ldots, DoFs}(\mathbf{q}_{f_i})\\
& d_{sph} \geq r_t + r_{f_i} - \epsilon\\
\label{eq:optProblem}
\end{aligned}
\end{equation}
where $r_t$ is the thumb radius at the fingertip (the half of the thumb width defined in Table \ref{tab:FingerProportions}), $r_{f_i}$ is the radius of the opposed finger at the desired point of interest (half of the joint or fingertip width defined in Table \ref{tab:FingerProportions}), $\epsilon$ is an arbitrary fixed scalar (e.g. 0.5 mm) which defines the intersection acceptance between the two chains, considering that finger surfaces can be made of soft materials, and $d_{sph}$ is the distance between two sphere centers, one located within the thumb and the other one within the opposed finger. Spheres are placed so that their surfaces overlap with the finger ones. Obviously, contour and constraining inequalities have to be adapted according to the finger shapes.

\subsection{Volume of intersection}
\label{subsec:volume of intersection}
\begin{figure}[tb]
\centering
\includegraphics[scale = 0.34]{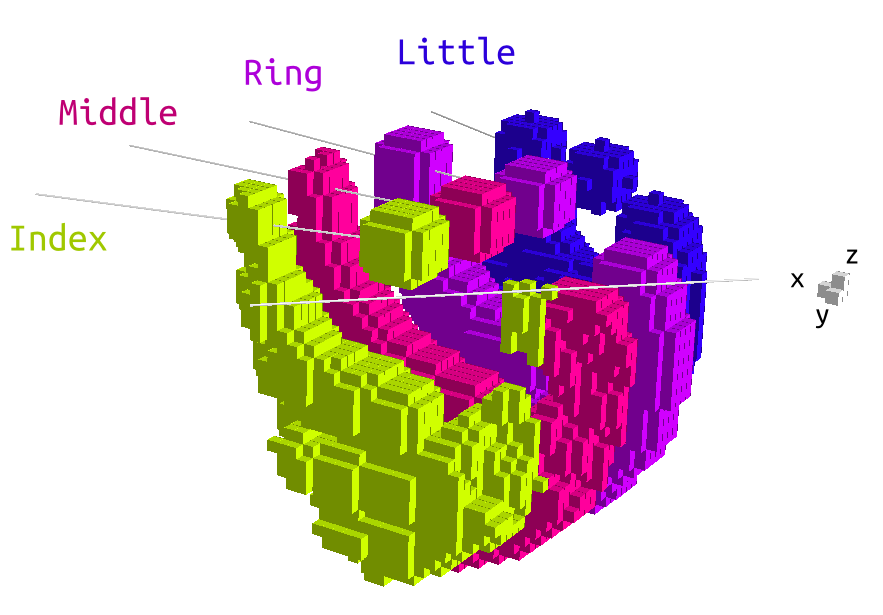}
\caption{Intersection between thumb fingertip and all fingers end-effectors.}
\label{fig:thumbFingerEeIntersection}
\end{figure}

Volumes of intersection among fingers discerns their kinematic degree of interaction. Collision points are identified by exploring the finger workspaces. To evaluate the thumb opposability a modified version of the performance index \cite{GifuIII} is used:
\begin{ceqn}
\begin{equation}
I = \frac{1}{d_t^3} \sum_{i = 1}^{k} \sum_{j = 1}^{e} w_{ij} v_{ij}
\label{eq:indexThumbOpposability}
\end{equation}
\end{ceqn}
where $d_t$ is the thumb length, $k$ is the number of fingers (thumb excluded), $e$ is the number of end-effectors considered on each opposed finger, $w_{ij}$ is a weighing coefficient, $v_{ij}$ is the volume of intersection between the thumb and the finger end-effector. This performance index includes and weights the interaction between the thumb fingertip and each finger portions involved in the Kapandji test. Intersection volumes ($v_{ij}$) are computed sampling the Cartesian space in $x$, $y$ and $z$. The number of cells which compose the grid strictly depends on the hand workspace boundaries and the size of the sampling interval (set to 2 mm). Each contact is checked by visiting the thumb and finger joint spaces and it is stored in the corresponding cell. Consequently, the number of cells holding collisions reveals the intersection volume per finger:
\begin{ceqn}
\begin{equation}
v_{ij} = n_{ij} \Delta V
\label{eq:intersectionVolue}
\end{equation}
\end{ceqn}
where $n_{ij}$ is the number of cells in which the thumb fingertip and the $j^{th}$ end-effector of the $i^{th}$ finger intersect and $\Delta V$ is the volume of a cell.\\
Differently from \cite{designGuidelineForTheThumb}, the end-effector is not considered as a single point. Indeed, a single point does not determine the potential grasping capabilities of the hand. For this reason, a cloud of points within the finger and thumb mechanical structures is taken into account. In this case, $v_{ij}$ approximatively discloses the amount of effective interaction between the thumb and the fingers close to the Kapandji positions (Figure \ref{fig:thumbFingerEeIntersection}).

\subsection{Solution selection}
\label{subsec:solution selection}
To select the best position and orientation of the first joint of the thumb the following algorithm is implemented:
\begin{enumerate}
\item Thumb bases are generated using a brute force exploration. Intervals are defined according to aesthetics and mechanical constraints;
\item For each base the Kapandji test is performed;
\item If the candidate passes the Kapandji test: the thumb opposability index is computed beside the relative standard deviation ($\sigma_r$) of the intersection volume:
\begin{ceqn}
\begin{equation}
\sigma_r = \frac{\sigma^2}{\bar{v}} 100
\label{eq:relStdDev}
\end{equation}
\end{ceqn}
where $\sigma^2 = \sum_{i = 1}^{k}(v_i - \bar{v})^2/ k$, $v_i = \sum_{j = 1}^{e} v_{ij}$ and $\bar{v}$ is the mean value of $v_i$.
\item If the relative standard deviation is lower than a desired threshold (in our case set to 20$\%$) the candidate is stored (the thumb opposability is enough equilibrated). 
\item The solution is selected picking the stored candidate with the highest thumb opposability index.
\end{enumerate}
Table \ref{tab:ThumbBaseplacement} lists some stored candidates. The one with the highest TOI is chosen.

\begin{table}[tb]
\centering
\caption{Candidate thumb bases. TOI is the Thumb Opposability Index}
\begin{tabular}{cccc}
\multicolumn{1}{c}{\multirow{1}{*}{\textbf{($x$, $y$, $z$) [mm] }}} & \multicolumn{1}{c}{\multirow{1}{*}{\textbf{orientation [deg]}}} & \multicolumn{1}{c}{\multirow{1}{*}{\textbf{TOI}}} & \multicolumn{1}{c}{\multirow{1}{*}{\textbf{$\mathbf{\sigma_r}$ [$ \%$]}}}\\
\midrule
\multicolumn{1}{c}{(8, 11, -9)} & \multicolumn{1}{c}{45$^{\circ}$} & \multicolumn{1}{c}{\multirow{1}{*}{0.167}} & \multicolumn{1}{c}{\multirow{1}{*}{19.693}}\\
\multicolumn{1}{c}{(8, 11, -5)} & \multicolumn{1}{c}{40$^{\circ}$} & \multicolumn{1}{c}{\multirow{1}{*}{0.169}} & \multicolumn{1}{c}{\multirow{1}{*}{17.421}}\\
\multicolumn{1}{c}{(8, 2, -5)} & \multicolumn{1}{c}{50$^{\circ}$} & \multicolumn{1}{c}{\multirow{1}{*}{0.177}} & \multicolumn{1}{c}{\multirow{1}{*}{6.705}}\\
\multicolumn{1}{c}{(12, 2, -5)} & \multicolumn{1}{c}{45.0$^{\circ}$} & \multicolumn{1}{c}{\multirow{1}{*}{0.196}} & \multicolumn{1}{c}{\multirow{1}{*}{12.252}}\\
\midrule
\end{tabular}
\label{tab:ThumbBaseplacement}
\end{table}

\section{Validation and discussion}
\label{sec:validation and discussion}
\begin{figure}[!t]
\centering
\hfill
\includegraphics[scale = 0.24]{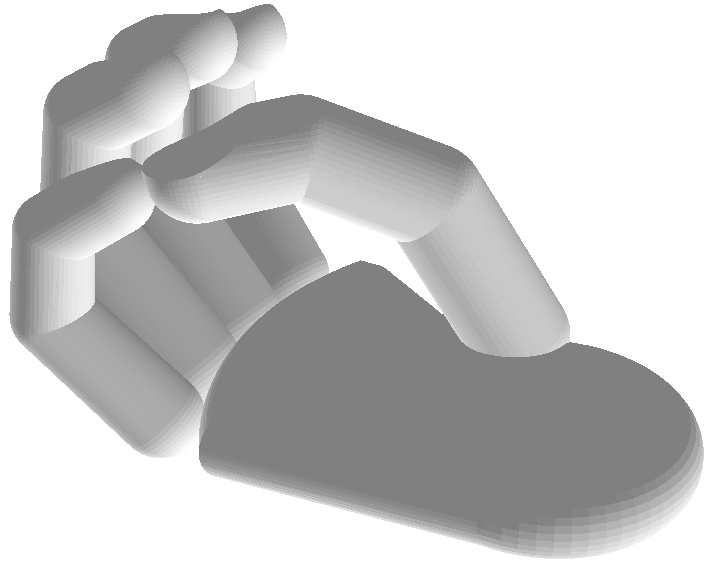}
\hfill(a)

\noindent\hfill
\includegraphics[scale = 0.22]{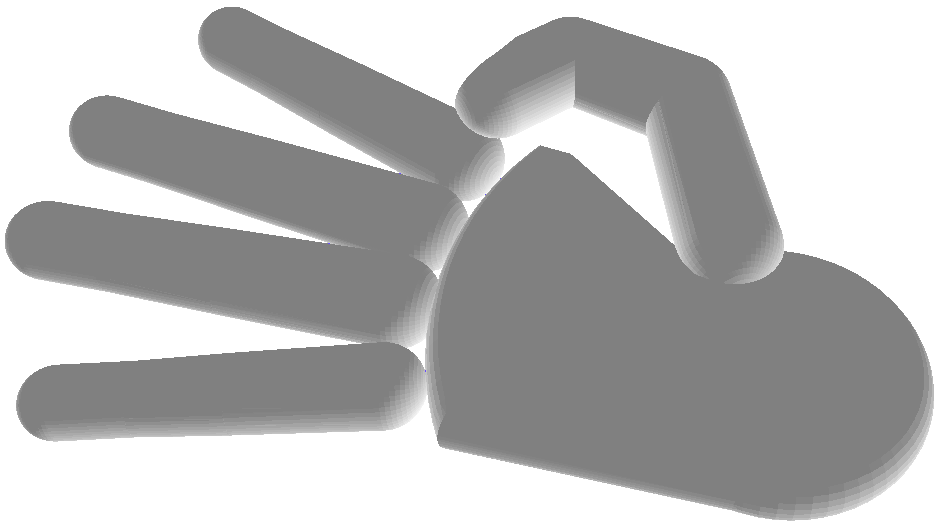}
\hfill(b)
\caption{Thumb opposition on the little fingertip (a) and the index MCP (b): two of the 16 positions demanded in the implemented Kapandji test. The thumb base is the last configuration listed in Table \ref{tab:ThumbBaseplacement}.}
\label{fig:LittleThumbFingertip}
\end{figure}

\begin{figure}[!t]
\centering
\hfill
\includegraphics[scale = 0.40]{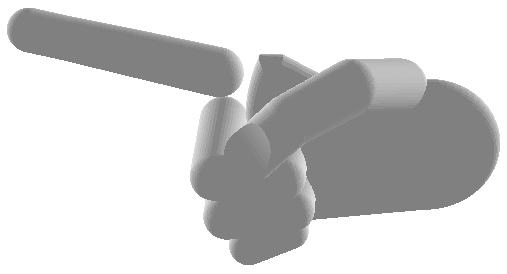}
\hfill(a)

\noindent\hfill
\includegraphics[scale = 0.28]{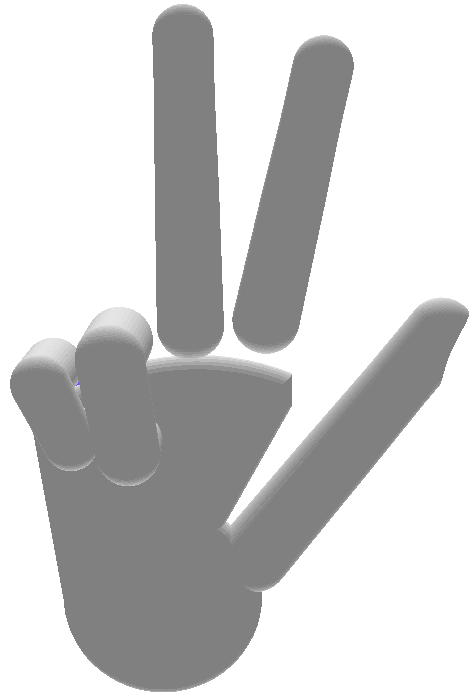}
\hfill(b)
\caption{Two examples of gestures that can be performed by the hand: pointing (a) and counting (b).}
\label{fig:hangGestures}
\end{figure}

The hand model is designed and simulated using Python and \textit{NAOqi}\cite{NAOqi} APIs. Geometric and kinematic parameters are inspected by direct visualization of the final robotic hand model. The Kapandji test is verified by moving the fingers and thumb in the collected configurations obtained during the algorithm execution and it is checked that the hand is able to reach the desired list of positions within the demanded precision. Figure \ref{fig:LittleThumbFingertip} shows the furthest (a) and the closest (b) positions involved in the test. Hand grasping capabilities have been examined on various objects (e.g. soda can and a smartphone) using CAD software. However, no conclusions can be drawn on the grasping success since only a kinematic analysis has been conducted at this stage. Beside common gestures, like "pointing" (Figure \ref{fig:hangGestures}-a), "thumb-up" or "ok" gestures, the hand is able to count from one to ten in the Chinese way and from one to five in the European one. For example, Figure \ref{fig:hangGestures}-b shows the number three in the European counting. Note that, in order to give more emphasis to the finger kinematics a simplified palm is represented. 

\section{Conclusions}
\label{sec:Conclusions}
A simple and effective numerical method for designing dexterous and highly opposable anthropomorphic robotic hands has been presented. The method is conceived to realize a fully embedded, small sized and light hand for a social humanoid robot of similar height to a 6 year old child. It consists in a design tool that generates the hand kinematics which are able to express emotions (gestures) and grasp small objects. Hand geometry has been conceived using anthropomorphic data and heuristics methods which allow a kinematic solution close to that of a human hand. The main focus has been given to the thumb base position and orientation, since its interaction with the opposed fingers strongly effects the hand performance. Three tests drove its selection: aesthetic, surgical and interaction tests. The first played a relevant role to obtain a fairly equilibrated hand, the second provided a human-like dexterous thumb, while the last one defined its opposability. Hand designers can follow this method to immediately obtain the hand kinematic model and structure. Future work will consist in improving the candidate selection among thumb bases using genetic algorithms rather than a brute force approach.

\addtolength{\textheight}{-8cm}
\bibliographystyle{IEEEtran}  
\bibliography{Bibliography}

\end{document}